\documentclass[10pt,twocolumn,letterpaper]{article}

\usepackage{cvpr}      %
\usepackage{multirow}   %
\usepackage{afterpage}

\definecolor{cvprblue}{rgb}{0.21,0.49,0.74}
\usepackage[pagebackref,breaklinks,colorlinks,allcolors=cvprblue]{hyperref}

\def\paperID{11129} %
\def\confName{CVPR}
\def\confYear{2025}

\title{GoalFlow: Goal-Driven Flow Matching for  Multimodal Trajectories Generation in End-to-End Autonomous Driving}

\author{
Zebin~Xing\textsuperscript{1,2}${^*}$, Xingyu~Zhang\textsuperscript{2}\thanks{Equal contribution.}, Yang Hu\textsuperscript{2}, Bo Jiang\textsuperscript{4,2}\\
Tong He\textsuperscript{5}, Qian Zhang\textsuperscript{2}, Xiaoxiao Long\textsuperscript{3}, Wei Yin\textsuperscript{2}\thanks{Corresponding author, project leader. Email: yvanwy@outlook.com} \\
\textsuperscript{1}School of Artificial Intelligence, University of Chinese Academy of Sciences \hspace{0.4cm} \textsuperscript{2}Horizon Robotics \\ 
\textsuperscript{3}Nanjing University \hspace{0.4cm}
\textsuperscript{4}Huazhong University of Science \& Technology
\hspace{0.4cm} 
\textsuperscript{5}Shanghai AI Laboratory
}

\begin{document}
\maketitle

\begin{abstract}

We propose GoalFlow, an end-to-end autonomous driving method for generating high-quality multimodal trajectories. In autonomous driving scenarios, there is rarely a single suitable trajectory. Recent methods have increasingly focused on modeling multimodal trajectory distributions. However, they suffer from trajectory selection complexity and reduced trajectory quality due to high trajectory divergence and inconsistencies between guidance and scene information. To address these issues, we introduce GoalFlow, a novel method that effectively constrains the generative process to produce high-quality, multimodal trajectories. To resolve the trajectory divergence problem inherent in diffusion-based methods, GoalFlow constrains the generated trajectories by introducing a goal point. GoalFlow establishes a novel scoring mechanism that selects the most appropriate goal point from the candidate points based on scene information.
Furthermore, GoalFlow employs an efficient generative method, Flow Matching, to generate multimodal trajectories, and incorporates a refined scoring mechanism to select the optimal trajectory from the candidates. Our experimental results, validated on the Navsim\cite{Dauner2024_navsim}, demonstrate that GoalFlow achieves state-of-the-art performance, delivering robust multimodal trajectories for autonomous driving. GoalFlow achieved \textbf{PDMS of 90.3}, significantly surpassing other methods. Compared with other diffusion-policy-based methods, our approach requires only \textbf{a single denoising step} to obtain excellent performance. The code is available at \url{https://github.com/YvanYin/GoalFlow}.

\end{abstract}
    
\section{Introduction}
\label{sec:intro}
\begin{figure}[h] %
    \centering
    \includegraphics[width=0.46\textwidth]{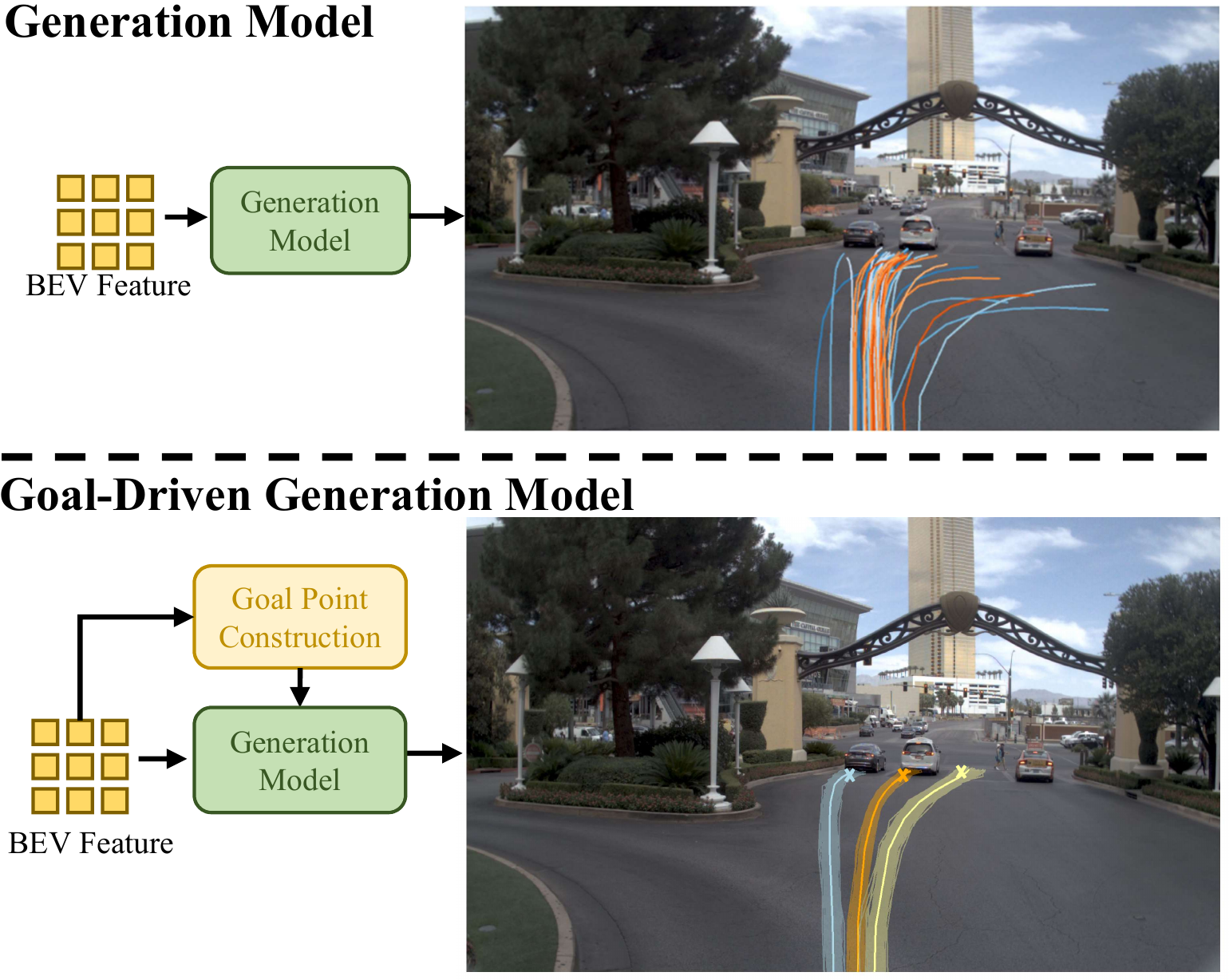}
    \caption{\textbf{The comparison of different multimodal trajectory generation paradigms recently.} A standalone generative model often produces highly diverse trajectories with no clear boundaries between different modalities. In contrast, the Goal-Driven Generation Model leverages the strong guidance of goal points, effectively distinguishing multiple modalities by utilizing different goal points.}
    \vspace{-20pt} 
    \label{fig:teaser}
\end{figure}

Since UniAD\cite{hu2023_uniad}, autonomous driving has increasingly favored end-to-end systems, where tasks like mapping and detection ultimately serve the planning task. To enhance system reliability, some end-to-end algorithms\cite{jiang2023vad,sun2024sparsedrive,huang2024gendrive} have begun exploring ways to generate multimodal trajectories as trajectory candidates for the algorithms. In autonomous driving, command typically includes indicators for left, right, and straight actions. VAD\cite{jiang2023vad} uses this command information to generate multimodal trajectories. Goal points, which provide the vehicle's location information for the next few seconds, are commonly used as guiding information in other approaches, such as SparseDrive\cite{sun2024sparsedrive}. These methods pre-define a set of goal points to generate different trajectory modes. Both approaches have succeeded in autonomous driving, offering candidate trajectories that significantly reduce collision rates. However, since these methods' guiding information does not pursue accuracy but instead provides a set of candidate values for the trajectory, when the gap between the guiding information and the ground truth is large, it is prone to generating low-quality trajectories.

In recent trajectory prediction works, some methods\cite{Jiang_2023_motiondiffuser, yang2024diffusiones, HE-drive} aim to generate multimodal trajectories through diffusion, using scene or motion information as a condition to produce multimodal trajectories. Other methods \cite{DOME} utilize diffusion to construct a world model. Without constraints, approaches like Diffusion-ES\cite{yang2024diffusiones} tend to generate divergent trajectories, which is depicted in the second row of Fig.\ref{fig:teaser}, requiring a scoring mechanism based on HD maps to align with the real-world road network, which is difficult to obtain in end-to-end environments. MotionDiffuser\cite{Jiang_2023_motiondiffuser} addresses trajectory divergence by using the ground truth endpoint as a constraint, which introduces overly strong prior information. GoalGAN\cite{dendorfer2020accv} first predicts the goal point and then uses it to guide the GAN network to generate trajectories. However, GoalGAN employs grid-cell to sample goal points, which does not consider the distribution of the goal points.

Reviewing previous work, we identified some overlooked issues:(1) Existing end-to-end autonomous driving systems tend to focus heavily on collision and L2 metrics, often adding specific losses or applying post-processing to reduce collision, while overlooking whether the vehicle remains within the drivable area. (2) Most end-to-end methods are based on regression models and aim to achieve multimodality by using different guiding information. However, when the guiding information deviates significantly from the ground truth, it can lead to the generation of low-quality trajectories. 

GoalFlow can be divided into three parts: Perception Module, Goal Point Construction Module, and Trajectory Planning Module. In the first module, following transfuser\cite{Chitta2023transfuser}, images and LiDAR are fed into two separate backbones and fused into BEV feature finally. In the second module, GoalFlow establishes a dense vocabulary of goal points, and a novel scoring mechanism is used to select the optimal goal point that is closest to the ground truth goal point and within a drivable area. In the third module, GoalFlow uses flow matching to model multimodal trajectories efficiently. It conditions scene information and incorporates stronger guidance from the selected goal point. Finally, GoalFlow employs a scoring mechanism to select the optimal trajectory. Compared to directly generating trajectories with diffusion, as in the first row of Fig. \ref{fig:teaser}, our approach provides strong constraints on the trajectory, leading to more reliable results.

We conducted experimental validation in Navsim and found that our method outperformed other approaches in overall scoring. Notably, due to our goal point selection mechanism, we achieved a significant improvement in DAC scores. Additionally, we observed that this flow-matching-based approach is robust to the number of denoising steps during inference. Even with only a single denoising step, the score dropped by only 1.6\% compared to the optimal case, enhancing the potential for real-world deployment of generative models in autonomous driving.

Our contributions can be summarized as follows:
\begin{itemize}
    \item We designed a novel approach to establishing goal points, demonstrating its effectiveness in guiding generative models for trajectory generation.
    \item We introduced flow matching to end-to-end autonomous driving and seamlessly integrated it with goal point guidance. 
    \item We developed an innovative trajectory selection mechanism, using shadow trajectories to further address potential goal point errors.
    \item Our method achieved state-of-the-art results in Navsim.
\end{itemize}

\section{Related Work}
\label{sec:related_work}

\begin{figure*}[ht]
  \centering
  \includegraphics[width=1.0\textwidth]{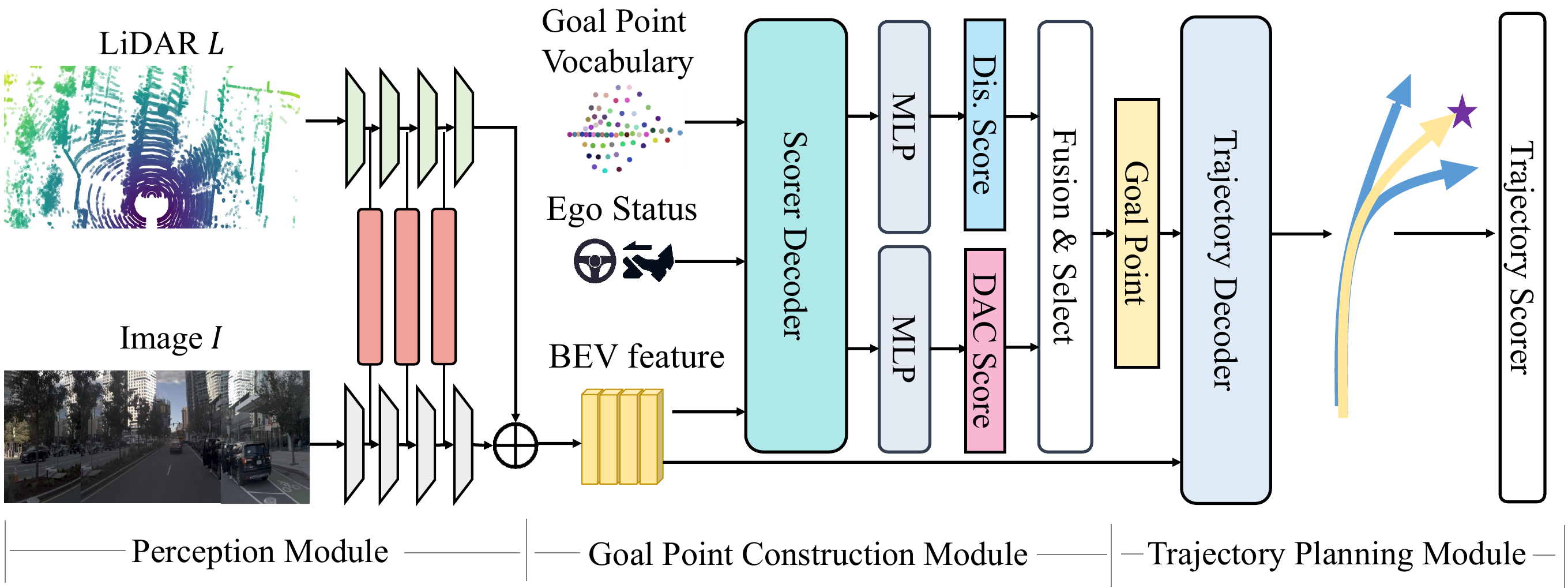}

  \caption{\textbf{Overview of the GoalFlow architecture.} GoalFlow consists of three modules. The Perception Module is responsible for integrating scene information into a BEV feature $F_{bev}$, the Goal Point Construction Module selects the optimal goal point from Goal Point Vocabulary $\mathbb{V}$ as guidance information, and the Trajectory Planning Module generates the trajectories by denoising from the Gaussian distribution to the target distribution. Finally, the Trajectory Scorer selects the optimal trajectory from the candidates.}
  \label{fig:main_fig}
\end{figure*}
\subsection{End-to-End Autonomous Driving}
Earlier end-to-end autonomous driving approaches\cite{Codevilla_imitation1}\cite{Codevilla_imitation2} used imitation learning methods, directly extracting features from input images to generate trajectories. Later, Transfuser\cite{Chitta2023transfuser} advanced by fusing lidar and image information during perception, using auxiliary tasks such as mapping and detection to provide supervision for the perception. FusionAD\cite{yetengju2023fusionad} took Transfuser a step further by propagating fused perception features directly to the prediction and planning modules. Other methods \cite{ADAPT,TOD3Cap} align the traffic scene with natural language. UniAD\cite{hu2023_uniad} introduced a unified query design that made the framework ultimately planning-oriented. Similarly, VAD\cite{jiang2023vad} focused on a planning-oriented approach by simplifying perception tasks and transforming scene representation into a vectorized format, significantly enhancing both planning capability and efficiency. Building on this, some methods\cite{chen2024vadv2,li2024hydra} discretized the trajectory space and constructed a trajectory vocabulary, transforming the regression task into a classification task. PARA-Drive\cite{weng2024_paradrive} performs mapping, planning, motion prediction, and occupancy prediction tasks in parallel. GenAD\cite{zheng2024genad} employed VAE and GRU for temporal trajectory reconstruction, while SparseDrive\cite{sun2024sparsedrive} progressed further in the vectorized scene representation, omitting denser BEV representations. Compared to previous methods that focus on better fitting ground truth trajectories using a regression model, we concentrate on generating high-quality multimodal trajectories in an end-to-end setting.

\subsection{Diffusion Model and Flow Matching}
Early generative models always used VAE\cite{kingma2013VAE} and GAN\cite{IanJ_GAN} in image generation. Recently, diffusion models that generate images by iteratively adding and removing noise have become mainstream. DDPM\cite{NEURIPS2020_ddpm} applies noise to images during training, converting states over time steps, and subsequently denoises them during testing to reconstruct the image. More recent methods\cite{song2020ddim} have further optimized sampling efficiency. Additionally, CFG\cite{Ho2022ClassifierFreeDG} has enhanced the robustness of generated outputs. Flow Matching\cite{DBLP:conf/iclr/flow_matching} establishes a vector field for transitioning from one distribution to another. Rectified flow\cite{DBLP:conf/iclr/rectifed_flow}, a specific form of flow matching, enables a direct, linear transition path between distributions. Compared to diffusion models, rectified flow often requires only a single inference step to achieve good results.

\subsection{MultiModal Trajectories Generation}
In planning tasks, such as manipulation and autonomous driving, a given scenario often offers multiple action options, requiring effective multimodal modeling. Recent works\cite{chi2023diffusionpolicy,pmlr-v229-chineddiffuser} in manipulation have explored this by applying diffusion models with notable success. Autonomous driving has adopted two main multimodal strategies: the first uses discrete commands to guide trajectory generation, such as in VAD\cite{jiang2023vad}, which produces three distinct trajectory modes, and SparseDrive\cite{sun2024sparsedrive} and \cite{huang2024gendrive}, which cluster fixed navigation points from datasets for trajectory guidance. The second approach introduces diffusion models directly to generate multimodal trajectories\cite{yang2024diffusiones,Jiang_2023_motiondiffuser,wang2024he}, achieving success in trajectory prediction but facing challenges in end-to-end applications. Building on diffusion models, we address limitations in accuracy and efficiency by incorporating flow matching, using goal points to guide trajectories with precision rather than focusing solely on multimodal diversity.

\section{Method}
\label{sec:method}

\subsection{Preliminary}
Compared to diffusion, which focuses on learning to reverse the gradual addition of noise over time to recover data, flow matching\cite{DBLP:conf/iclr/flow_matching} focuses on learning invertible transformations that map between data distributions.
Let $\pi_0$ denote a simple distribution, typically the standard normal distribution $p(x)=\mathcal{N}(x|0,I)$, and let $\pi_1$ denote the target distribution. Under this framework, rectified flow\cite{DBLP:conf/iclr/rectifed_flow} uses a simple and effective method to construct the path through optimal transport\cite{OT_displacement} displacement, which we choose as our Flow Matching method.

Given $x_0$ sampled from $\pi_0$, $x_1$ sampled from $\pi_1$, and $t\in[0,1]$,  the path from $x_0$ to $x_1$ is defined as a straight line, meaning the intermediate status $x_t$ is given by $(1-t)x_0+tx_1$, with the direction of intermediate status consistently following $x_1-x_0$. By constructing a neural network $v_{\theta}$ to predict the direction $x_1-x_0$ based on the current state $x_t$ and time step $t$, we can obtain a path from the initial distribution $\pi_0$ to target distribution $\pi_1$ by optimizing the loss between $v_\theta(x_t,t)$ and $x_1-x_0$. This can be formalized as:
\begin{equation}
    v_{\theta}(x_t,t) \approx \mathbf{E}_{x_0\sim\pi_0,x_1\sim\pi_1}[v_t|x_t]
    \label{eq:rectified_flow}
\end{equation}
\begin{equation}
    \mathcal{L}(\theta)=\mathbf{E}_{x_0\sim\pi_0,x_1\sim\pi_1}[\| v_\theta(x_t, t) - (x_1 - x_0) \|_2 ]
    \label{eq:rectified_flow_loss}
\end{equation}

\subsection{GoalFlow}
\subsubsection{Overview}
GoalFlow is a goal-driven end-to-end autonomous driving method that can generate high-quality multimodal trajectories. The overall architecture of GoalFlow is illustrated in Figure \ref{fig:main_fig}. It comprises three main components. In the Perception Module, we obtain a BEV feature $F_{\text{bev}}$ that encapsulates environmental information by fusing camera images $I$, and LiDAR data $L$. The Goal Point Construction Module focuses on generating precise guidance information for trajectory generation. It accomplishes this by constructing a goal point vocabulary $\mathbb{V}=\{g_i\}^N$, and employing a scoring mechanism to select the most appropriate goal point $g$. In the Trajectory Planning Module, we produce a set of multimodal trajectories, $\mathbb{T} = \{\hat{\tau}_i\}^M$, and then identify the optimal trajectory $\tau$, through a trajectory scoring mechanism.

\subsubsection{Perception Module}
In the first step, we fuse image and LiDAR data to create a BEV feature, $F_{\text{bev}}$, that captures rich road condition information. A single modality often lacks crucial details; for example, LiDAR does not capture traffic light information, while images cannot precisely locate objects. By fusing different sensor modalities, we can achieve a more complete and accurate representation of the road conditions.

We adopt the Transfuser architecture \cite{Chitta2023transfuser} for modality fusion. The forward, left, and right camera views are concatenated into a single image $I \in \mathbb{R}^{3 \times H_1 \times W_1}$, while LiDAR data is formed as a tensor $L \in \mathbb{R}^{K \times 3}$. These inputs are passed through separate backbones, and their features are fused at different layers using multiple transformer blocks. The result is a BEV feature, $F_{\text{bev}}$, which comprehensively represents the scene. To ensure effective interaction between the ego vehicle and surrounding objects, as well as map information, we apply auxiliary supervision to the BEV feature through losses derived from HD maps and bounding boxes.

\subsubsection{Goal Point Construction Module.}
In this module, we construct a precise goal point to guide the trajectory generation process. Diffusion-based approach\cite{yang2024diffusiones,Jiang_2023_motiondiffuser} without constraints often leads to excessive trajectory divergence, which complicates trajectory selection. Our key observation is that a goal point contains a precise description of the short-term future position, which imposes a strong constraint on the generation model. As a result, we divide the traditional Planning Module into two steps: first, constructing a precise goal point, and second, generating the trajectory through planning.

\textbf{Goal Point Vocabulary.}
We aim to construct a goal point set that provides candidates for the optimal goal point. Traditional goal-based methods\cite{densetnt,pmlr-tnt}, rely on lane-level information from HD map to generate goal point sets for trajectory prediction. However, HD maps are expensive, making lane information often unavailable in end-to-end driving. Inspired by VADv2\cite{chen2024vadv2}, we discretize the endpoint space of trajectories to generate candidate goal points, enabling a solution without relying on HD maps. We clustered trajectory endpoints $\mathbf{p}_i = (x_i, y_i, \theta_i)$ in the training data to create $N$ cluster centers, which form our goal point vocabulary
$\mathbb{V}$. Each endpoint $p_i$ represents a position $(x_i,y_i)$ and heading $\theta_i$. To ensure that the vocabulary represents finer-grained locations, we typically set $N$ to a large value, generally 4096 or 8192.

\begin{figure*}[htbp]
    \centering
    \begin{minipage}{0.55\textwidth}
        \centering
        \includegraphics[height=6.0cm]{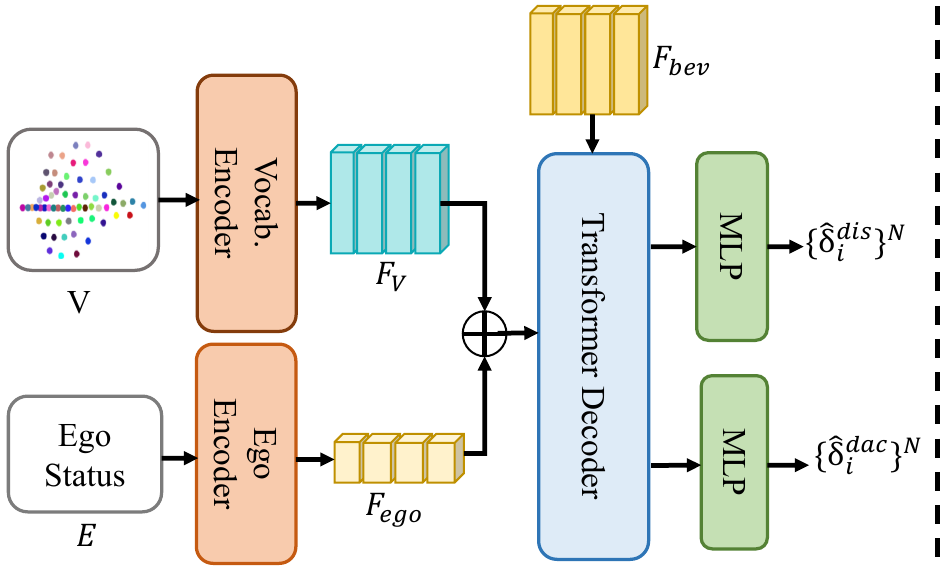} %
        \caption*{(a)} %
        \label{fig:image1}
    \end{minipage}%
    \begin{minipage}{0.45\textwidth}
        \centering
        \includegraphics[height=6.0cm]{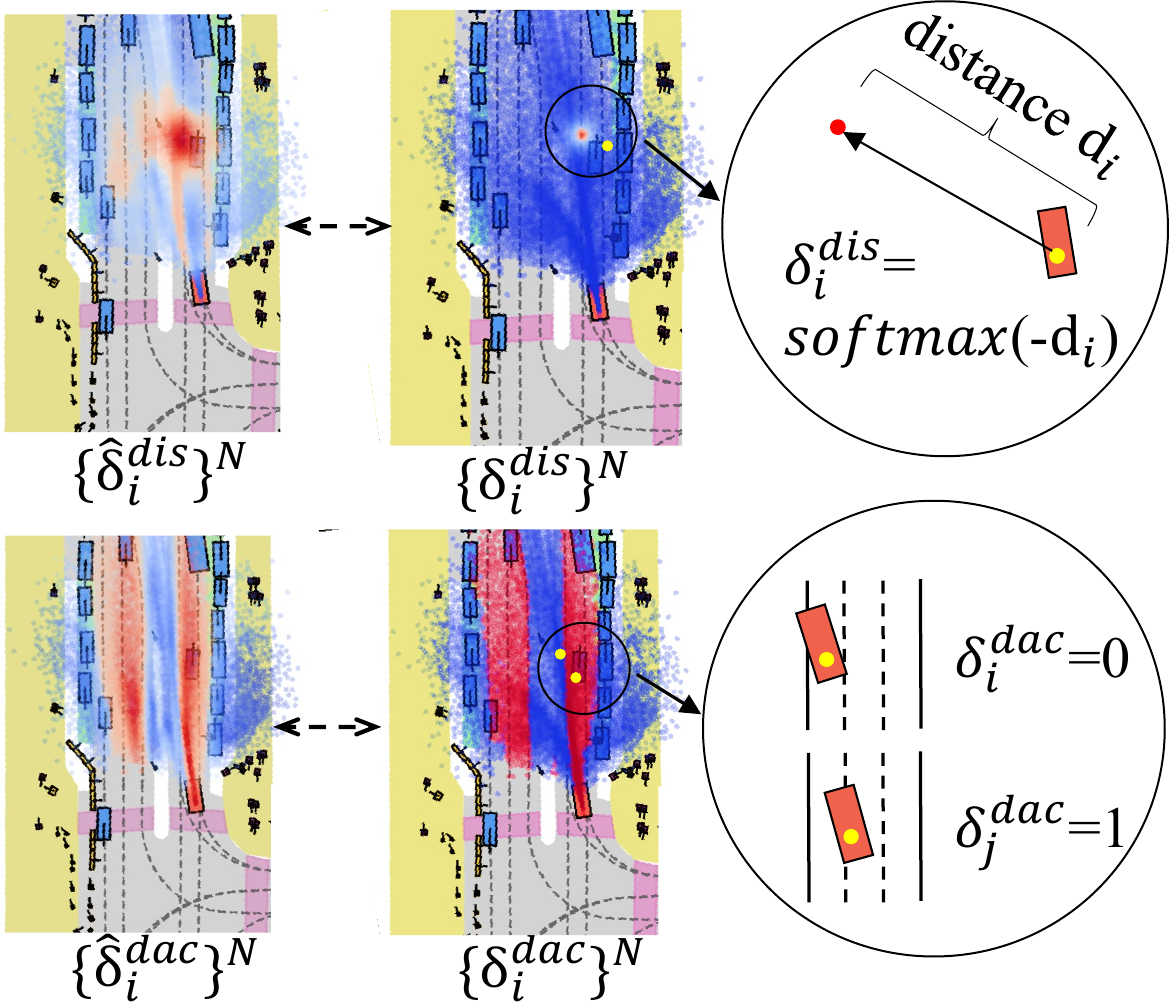} %
        \caption*{(b)} %
        \label{fig:image2}
    \end{minipage}
    \caption{\textbf{Goal Point Scorer.} (a) shows the detailed structure of the Goal Point Construction Module, and (b) presents the score distributions of $\{\hat{\delta}^{dis}_i\}^N$, $\{\hat{\delta}^{dac}_i\}^N$, and $\{\hat{\delta}^{final}_i\}^N$, where points with higher scores are highlighted with warmer color.}
    \label{fig:overall}
\end{figure*}

\begin{figure}[ht] %
    \centering
    \includegraphics[width=0.5\textwidth]{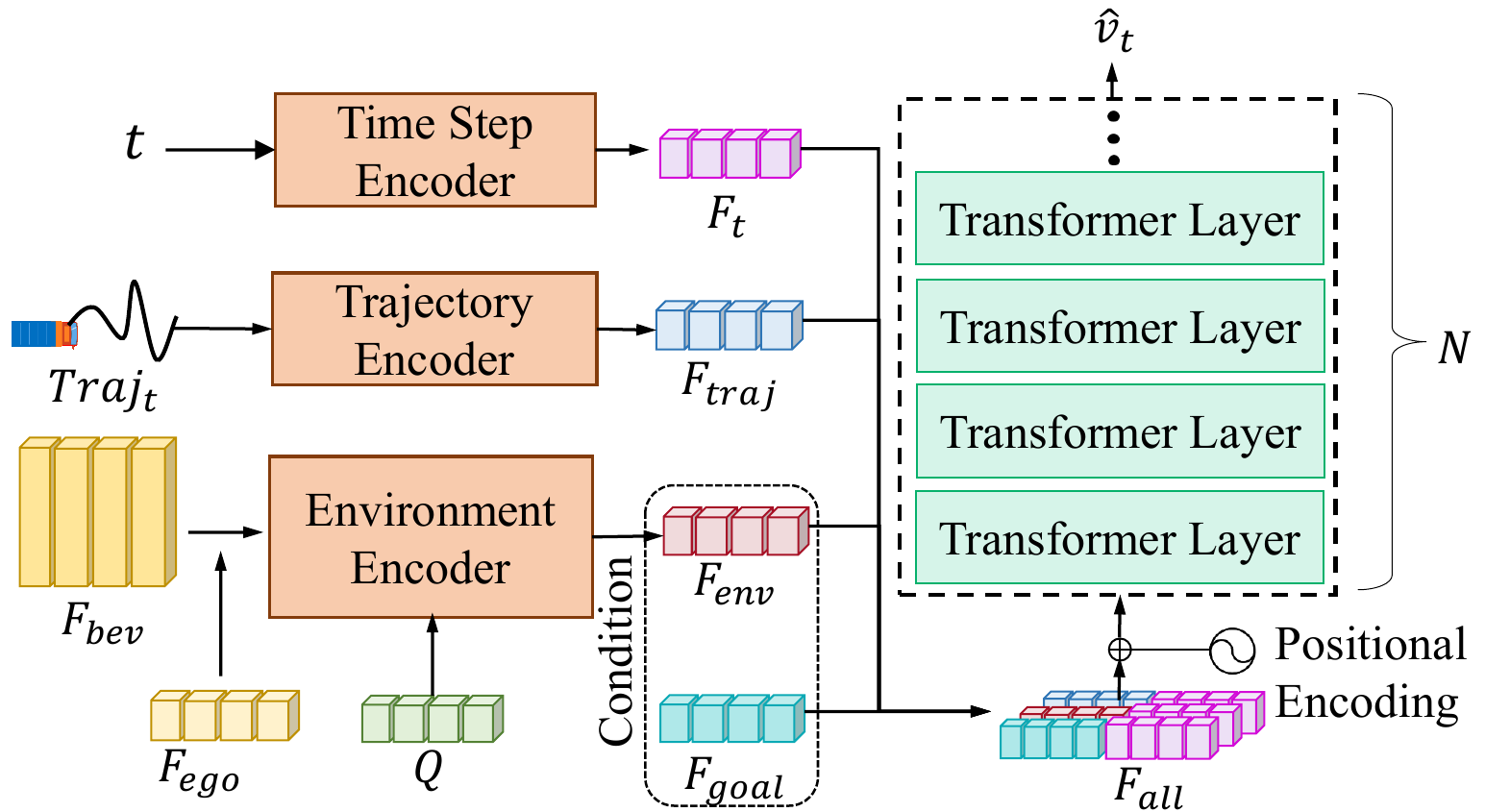}
    \caption{\textbf{The network architecture used in Rectified Flow.} 
    }
    \label{fig:rectified_flow}
\end{figure}

\textbf{Goal Point Scorer.}
High-quality trajectories typically exhibit the following characteristics: A small distance to the ground truth and within the drivable area. To achieve this, we evaluate each goal point \( g_i \) in the vocabulary \( \mathbb{V} \) using two distinct scores: the \textit{Distance Score} \( \hat{\delta}^{\text{dis}} \) and the \textit{Drivable Area Compliance Score} \( \hat{\delta}^{\text{dac}} \). The Distance Score measures the proximity between the goal point \( g_i \) and the endpoint of ground truth trajectory \( g^{\text{gt}} \), with a continuous value in the range \( \hat{\delta}^{\text{dis}} \in [0, 1] \), where a higher value indicates a closer match to \( g^{\text{gt}} \). The Drivable Area Compliance Score ensures that the goal point lies within the drivable area, using a binary value \( \hat{\delta}^{\text{dac}} \in \{0, 1\} \), where 1 indicates that the goal point is valid within the drivable area, and 0 indicates it is not. 

To construct the target distance score \( \delta^{\text{dis}}_i \), we utilize the softmax function to map the Euclidean distance between the goal point \( g_i \) and the ground truth goal point \( g^{\text{gt}} \) to the interval \([0, 1]\). This is defined as:

\begin{equation}
    \delta^{\text{dis}}_i = \frac{\exp(-\|g_i - g^{\text{gt}}\|_2)}{\sum_{j} \exp(-\|g_j - g^{\text{gt}}\|_2)}
    \label{eq:distance score}
\end{equation}

For the target drivable area compliance score \( \delta^{\text{dac}}_i \), we introduce a shadow vehicle, whose bounding box is determined based on the position and heading \((x_i, y_i, \theta_i)\) in \( g_i \) and the shape of the ego vehicle. Let \( \{p^j\}^4 \) represent the set of four corner positions of the shadow vehicle, and let $\mathbb{D}$ denote the polygon representing the drivable area. The drivable area compliance score \( \delta^{\text{dac}}_i \) is defined as:

\[
\delta^{\text{dac}}_i =
\begin{cases}
1, & \text{if } \forall j, \, p^j \in \mathbb{D}^\circ \\
0, & \text{otherwise}
\end{cases}
\]
\label{eq:dac score}

We compute the final score \( \hat{\delta}^{\text{final}}_i \) by aggregating \( \hat{\delta}^{\text{dis}}_i \) and \( \hat{\delta}^{\text{dac}}_i \). The goal point with the highest final score is selected for trajectory generation.

\[
\hat{\delta}^{\text{final}}_i = w_1 \log \hat{\delta}^{\text{dis}}_i + w_2 \log \hat{\delta}^{\text{dac}}_i
\]
\label{eq:target_score}

As shown in Fig.\ref{fig:overall}(a), the Transformer-based Scorer Decoder uses the result of adding $F_v$ and $F_{\text{ego}}$ as the query, with $F_{\text{bev}}$ as the key and value. The output is passed through two separate MLPs to produce the scores $\hat{\delta}^{dis}$ and $\hat{\delta}^{dac}$ for each point in the $\mathbb{V}$. Fig.\ref{fig:overall}(b) shows the distribution of these two scores. With the points in warmer colors representing higher scores, we observe that score $\hat{\delta}^{dis}$ effectively indicates the desired future position, while $\hat{\delta}^{dac}$ identifies if the goal point is within the drivable area.

\subsubsection{Trajectory Planning Module}
In this module, we generate constrained, high-quality trajectory candidates using a generative model and then select the optimal trajectory through a scoring mechanism. Generative models based on diffusion methods like DDPM\cite{NEURIPS2020_ddpm} and DDIM\cite{song2020ddim} typically require complex denoising paths, leading to significant time overhead during inference, which makes them unsuitable for real-time systems like autonomous driving. In contrast, Rectified Flow\cite{DBLP:conf/iclr/rectifed_flow}, which is based on the optimal transport path in flow matching, requires much fewer inference steps to achieve good results. We adopt Rectified Flow as the generative model, using the BEV feature and goal point as conditions to generate multimodal trajectories.

\textbf{Multimodal Trajectories Generating.}
We generate multimodal trajectories by modeling the shift from the noise distribution to the target trajectory distribution. During this distribution transfer process, given the current state $x_t$ and time step $t$, we predict the shift $\mathbf{v_t}$.

\begin{equation}
    \mathbf{v_t}=\tau^{norm}-x_0
    \label{eq:rectified_flow_train3}
\end{equation}
\begin{equation}
    x_t=(1-t)x_0+t\tau^{norm}
    \label{eq:rectified_flow_train1}
\end{equation}
\begin{equation}
    \tau^{norm}=\mathcal{H}(\tau^{gt})
    \label{eq:rectified_flow_train1}
\end{equation}
Where $\tau^{gt}$ is the ground truth trajectory and $\tau^{norm}$ is its normalized form. We define $\mathcal{H}(\cdot)$ as the normalization operation applied to the trajectory. The variable $x_0$ represents the noise distribution, which follows $x_0 \sim \mathcal{N}(0, \sigma^2 I)$. The variable $x_t$ is obtained by linearly interpolating between $x_0$ and $\tau^{norm}$.

As illustrated in Fig.\ref{fig:rectified_flow}, we extract different features through a series of encoders. Specifically, we encode $x_t$ using a linear layer, while $t$ and the goal point are transformed into feature vectors via sinusoidal encoding. The feature $F_{\text{env}}$ is obtained by passing the information from $F_{\text{bev}}$ and $F_{\text{ego}}$ through the environment encoder.

\begin{equation}
    F_{\text{env}} = E_{\text{env}}(Q, (F_{\text{BEV}}+F_{\text{ego}}), (F_{\text{BEV}}+F_{\text{ego}}))
    \label{eq:env condition}
\end{equation}
Here, $E_{\text{env}}$ refers to a Transformer-based encoder, $Q$ denotes a learnable embedding, and $F_{\text{ego}}$ represents the ego status feature, which encodes the kinematic information of the ego vehicle.

We concatenate the features $F_{\text{env}}$, $F_{\text{goal}}$, $F_{\text{traj}}$, and $F_{\text{t}}$ to form the overall feature $F_{\text{all}}$, which encapsulates the current state, time step, and scene information. This combined feature is then passed through several attention layers to predict the distribution shift $\mathbf{v_t}$.

\begin{equation}
    \mathbf{\hat{v}_t}=\mathcal{G}(F_{\text{all}},F_{\text{all}},F_{\text{all}})
    \label{eq:env condition}
\end{equation}
\begin{equation}
    F_{\text{all}} = \text{Concat}(F_{\text{env}},F_{\text{goal}},F_{\text{traj}},F_t)
    \label{eq:env condition}
\end{equation}
Where $\mathcal{G}$ is the network that consists of N attention layers.

We reconstruct the trajectory distribution using $x_0$ and $\mathbf{\hat{v}_t}$. Typically, we achieve this by performing multiple inference steps through the Rectified Flow, gradually transforming the noise distribution $x_0$ to the target distribution $\tau^{\text{norm}}$. Finally, we apply denormalization to $\tau^{\text{norm}}$ to obtain the final trajectory $\hat{\tau}$.
\begin{equation}
    \hat{\tau}=\mathcal{H}^{-1}(\hat{\tau}^{norm})
    \label{eq:rectified_flow_infer1}
\end{equation}
\vspace{-1em}  %
\begin{equation}
    \hat{\tau}^{norm}=x_0+\frac{1}{n}\sum_{i}^{n}\hat{v}_{t_{i}}
    \label{eq:rectified_flow_infer2}
\end{equation}

Where $n$ is the total inference steps, and $t_i$ is the time step sampled in the $i$-th step, which satisfies $t_i\in[0,1]$. $\mathcal{H}^{-1}(\cdot)$ is the denormalization operation.

\textbf{Trajectory Selecting}
In trajectory selection, methods like SparseDrive\cite{sun2024sparsedrive} and Diffusion-ES\cite{yang2024diffusiones} rely on kinematic simulation of the generated trajectories to predict potential collisions with surrounding agents, thus selecting the optimal trajectory. This process significantly increases the inference time. We simplify this procedure by using the goal point as a reference for selecting the trajectory. Specifically, we trade off the trajectory distance to the goal point and ego progress, selecting the optimal trajectory through a trajectory scorer.
\begin{equation}
    f(\hat{\tau}_i) = -\lambda_1{\Phi(f_{dis}(\hat{\tau}_i))}+\lambda_2{\Phi(f_{pg}(\hat{\tau}_i))}
    \label{eq:trajectory_score}
\end{equation}
where $\Phi(\cdot)$ is the minimax operation. $f_{dis}(\hat{\tau}_i)$ presents the $\mathcal{L}_2$ distance of $\hat{\tau}_i$ and $g$, and $f_{pg}(\hat{\tau}_i)$ presents the $\mathcal{L}_2$ distance of progress of $\hat{\tau}_i$ make.

Furthermore, predicted goal point may contain an error that can misguide the trajectory. To mitigate this, we mask the goal point during generation to create a shadow trajectory. If the shadow trajectory deviates significantly from the main trajectory, we treat the goal point as unreliable and use the shadow as the output.

\subsubsection{Training Losses}
Firstly, we
optimize the perception extractor exclusively, and enforce multiple perception losses for supervision, including the cross-entropy loss for HD map ($L_{HD}$) and 3D bounding box classification ($L_{bbox}$) and $L_1$ loss for 3D bounding box locations ($L_{loc}$). This stage aims to enrich the BEV feature with information on various perceptions. Losses are as follows.
\begin{equation}
    L_{perception} = w_1 * L_{HD} + w_2 * L_{bbox} + w_3 * L_{loc}
\end{equation}
where $w_1, w_2, w_3$ are set to 10.0, 1.0, 10.0 in training.
For the goal constructor, we employ the cross entropy loss for distance score($L_{dis}$) and DAC score($L_{dac}$). $w_4$, $w_5$ are set to 1.0 and 0.005.
\begin{equation}
    L_{goal} =w_4 * L_{dis} + w_5 * L_{dac}
\end{equation}
\begin{equation}
    L_{dis}=-\sum_{i=1}^{N}\delta_i^{dis} log(\hat{\delta_i}^{dis})
\end{equation}
\begin{equation}
    L_{dac}=-\delta^{dac}log\hat{\delta}^{dac}-(1-\delta^{dac})log(1-\hat{\delta}^{dac})
\end{equation}
$L_1$ loss is utilized for multimodal planner.
\begin{equation}
    L_{planner}=|\mathbf{v_t} - \mathbf{\hat{v}_t}|
\end{equation}

\section{Experiments}
\label{sec:experiments}

\begin{table*}[ht] %
    \centering
    \resizebox{\textwidth}{!}{
        \begin{tabular}{lcccccccccc}
            \toprule
            \textbf{Method} & \textbf{Ego Stat.} & \textbf{Image} & \textbf{LiDAR} & \textbf{Video} & $\mathbf{S_{NC}}$ $\uparrow$ & $\mathbf{S_{DAC}}$ $\uparrow$ & $\mathbf{S_{TTC}}$ $\uparrow$ & $\mathbf{S_{CF}}$ $\uparrow$ & $\mathbf{S_{EP}}$ $\uparrow$ & $\mathbf{S_{PDM}}$ $\uparrow$ \\
            \midrule
            Constant Velocity & \checkmark & & & & 68.0 & 57.8 & 50.0 & 100 & 19.4 & 20.6 \\
            Ego Status MLP & \checkmark & & & & 93.0 & 77.3 & 83.6 & 100 & 62.8 & 65.6 \\
            \midrule
            LTF \cite{Chitta2023transfuser} & \checkmark & \checkmark & & & 97.4 & 92.8 & 92.4 & 100 & 79.0 & 83.8 \\
            TransFuser \cite{Chitta2023transfuser} & \checkmark & \checkmark & \checkmark & & 97.7 & 92.8 & 92.8 & 100 & 79.2 & 84.0 \\
            UniAD \cite{hu2023_uniad} & \checkmark & \checkmark & \checkmark & & 97.8 & 91.9 & 92.9 & 100 & 78.8 & 83.4 \\
            PARA-Drive \cite{weng2024_paradrive} & \checkmark & \checkmark & \checkmark & \checkmark & 97.9 & 92.4 & 93.0 & 99.8 & 79.3 & 84.0 \\
            \textbf{GoalFlow (Ours)} & \checkmark & \checkmark & \checkmark & & \textbf{98.4} & \textbf{98.3} & \textbf{94.6} & \textbf{100} & \textbf{85.0} & \textbf{90.3} \\
            \midrule
            \textcolor{gray}{$\text{GoalFlow}^{\dag}$} & \textcolor{gray}{\checkmark} & \textcolor{gray}{\checkmark} & \textcolor{gray}{\checkmark} &  & \textcolor{gray}{99.8} & \textcolor{gray}{97.9} & \textcolor{gray}{98.6} & \textcolor{gray}{100} & \textcolor{gray}{85.4} & \textcolor{gray}{92.1} \\
            \textcolor{gray}{$\text{Human}^{\ddag}$} & & & &  & \textcolor{gray}{100} & \textcolor{gray}{100} & \textcolor{gray}{100} & \textcolor{gray}{99.9} & \textcolor{gray}{87.5} & \textcolor{gray}{94.8} \\
            \bottomrule
        \end{tabular}
    }
    \caption{\textbf{Comparisons with SOTA methods in PDM score metrics on Navsim \cite{Dauner2024_navsim} Test.} Our method outperforms other approaches across all evaluation metrics. $\dag$ uses the endpoint of the ground-truth trajectory as the goal point. $\ddag$ uses the ground-truth trajectories to evaluate.}
    \label{tab:main_table}
\end{table*}

\begin{table*}[ht] %
    \centering
    \begin{tabular}{l lcccccc} %
        \toprule
        \textbf{Model} & \textbf{Description} & $\mathbf{S_{NC}}$ $\uparrow$ & $\mathbf{S_{DAC}}$ $\uparrow$ & $\mathbf{S_{TTC}}$ $\uparrow$ & $\mathbf{S_{CF}}$& $\mathbf{S_{EP}}$ $\uparrow$ & $\mathbf{S_{PDM}}$ $\uparrow$ \\
        \midrule
        \textcolor{gray}{$-$} & \textcolor{gray}{Transfuser\cite{Chitta2023transfuser}} & \textcolor{gray}{97.7} & \textcolor{gray}{92.8} & \textcolor{gray}{92.8} & \textcolor{gray}{100} & \textcolor{gray}{79.0} & \textcolor{gray}{84.0} \\
        \midrule
        $\mathcal{M}_0$ & Base Model & 97.9 & 94.2 & 94.2 & 100 & 79.9 & 85.6 \\
        $\mathcal{M}_1$ & $\mathcal{M}_0$ + Distance Score Map & 98.5 & 96.4 & 94.9 & 100 & 83.0 & 88.5 \\
        $\mathcal{M}_2$ & $\mathcal{M}_1$ + DAC Score Map & \textbf{98.6} & 97.5 & \textbf{94.7} & \textbf{100} & 83.8 & 89.4\\
        $\mathcal{M}_3$ & $\mathcal{M}_2$ + Trajectory Scorer & 98.4 & \textbf{98.3} & 94.6 & 100 & \textbf{85.0} & \textbf{90.3} \\
        \bottomrule
    \end{tabular}
    \caption{\textbf{Ablation study on the influence of each component. }$\mathcal{M}_0$ is the base model, which uses rectified flow without goal point guidance and averages all generated trajectories to produce the final output. $\mathcal{M}_1$ and $\mathcal{M}_2$ introduce the distance score map and DAC score map, respectively, to guide the rectified flow. $\mathcal{M}_3$ builds upon $\mathcal{M}_1$ by incorporating trajectory scorer.}
    \label{tab:ablation_modules}
\end{table*}

\begin{table}[ht]
    \centering
    \resizebox{\linewidth}{!}{
        \begin{tabular}{lccccccc}
            \toprule
            \textbf{$T$} & $\mathbf{Inf. Time}$ & $\mathbf{S_{NC}}$ $\uparrow$ & $\mathbf{S_{DAC}}$ $\uparrow$ & $\mathbf{S_{TTC}}$ $\uparrow$ & $\mathbf{S_{CF}}$ $\uparrow$ & $\mathbf{S_{EP}}$ $\uparrow$ & $\mathbf{S_{PDM}}$ $\uparrow$ \\
            \midrule
            20 & 177.8ms & 98.3 & 98.1 & 94.3 & 100 & 84.7 & 89.9 \\
            10 & 92.4ms & 98.3 & 98.2 & 94.4 & 100 & \textbf{84.9} & 90.1 \\
            5 & 49.0ms & 98.4 & \textbf{98.3} & \textbf{94.6} & \textbf{100} & 84.4 & \textbf{90.3} \\
            1 & 10.4ms & \textbf{98.4} & 97.8 & 94.1 & 100 & 84.5 & 88.9 \\
            \bottomrule
        \end{tabular}
    }
    \caption{\textbf{Impact of different timesteps in inference.} $T$ denotes the number of denoising steps during inference. The results indicate that the model's performance is robust to variations of denoising steps.}
    \label{tab:ablation_steps}
\end{table}

\begin{table}[ht]
    \centering
    \resizebox{\linewidth}{!}{
        \begin{tabular}{lcccccc}
            \toprule
            \textbf{$\sigma$} & $\mathbf{S_{NC}}$ $\uparrow$ & $\mathbf{S_{DAC}}$ $\uparrow$ & $\mathbf{S_{TTC}}$ $\uparrow$ & $\mathbf{S_{CF}}$ $\uparrow$ & $\mathbf{S_{EP}}$ $\uparrow$ & $\mathbf{S_{PDM}}$ $\uparrow$ \\
            \midrule
            0.05 & 98.3 & 98.2 & 94.4 & 100 & \textbf{85.0} & 90.1 \\
            0.1 & \textbf{98.4} & \textbf{98.3} & \textbf{94.6} & \textbf{100} & 85.0 & \textbf{90.3} \\
            0.2 & 87.4 & 76.0 & 69.4 & 32.0 & 56.2 & 49.0 \\
            0.3 & 68.3 & 48.1 & 44.8 & 2.23 & 23.6 & 18.8 \\
            \bottomrule
        \end{tabular}
    }
    \caption{\textbf{Impact of different values of $\sigma$ on the initial noise distribution.} $\sigma$ is the standard deviation of $x_0$. The results show that performance drops significantly when $\sigma$ exceeds 0.1, but remains stable for values below 0.1.}
    \label{tab:noise_scale}
\end{table}

\subsection{Dataset}
Our experiment is validated on the Openscene\cite{openscene2023} dataset. Openscene includes 120 hours of autonomous driving data. Its end-to-end environment Navsim\cite{Dauner2024_navsim} uses 1192 and 136 scenarios for trainval and testing, a total of over 10w samples at 2Hz. Each sample contains camera images from 8 perspectives, fused Lidar data from 5 sensors, ego status, and annotations for the map and objects. 

\subsection{Metrics}
In the Navsim environment, the generated 2Hz, 4-second trajectories are interpolated via an LQR controller to yield 10Hz, 4-second trajectories. These trajectories are scored using closed-loop metrics, including No at-fault Collisions $S_{NC}$, Drivable Area Compliance $S_{DAC}$, Time to Collision $S_{TTC}$ with bounds, Ego Progress $S_{EP}$, Comfort $S_{CF}$, and Driving Direction Compliance $S_{DDC}$. The final score is derived by aggregating these metrics. Due to practical constraints, $S_{DDC}$ is omitted from the calculation\footnote{https://github.com/autonomousvision/navsim/issues/14}. 
\begin{equation}
\begin{aligned}
    S_{PDM} = &\ S_{NC} \times S_{DAC} \times s_{TTC} \times \\
           &\ \left( \frac{5 \times S_{EP} + 5 \times S_{CF} + 2 \times S_{DDC}}{12} \right)
\end{aligned}
\label{eq:pdm_scorer}
\end{equation}

\subsection{Baselines}
In Navsim, we compare against the following baselines:
\textbf{Constant Velocity}
Assumes constant speed from the current timestamp for forward movement.
\textbf{Ego Status MLP}
Takes only the current state as input and uses an MLP to generate the trajectory.
\textbf{PDM-Closed}
Using ground-truth perception as input, several trajectories are generated through a rule-based IDM method. The PDM scorer then selects the optimal trajectory from these as the output.
\textbf{Transfuser}
Uses both image and LiDAR inputs, fusing them via a transformer into a BEV feature, which is then used for trajectory generation.
\textbf{LTF}
A streamlined version of Transfuser, where the LiDAR backbone is replaced with a learnable embedding. It achieves results in NavSim similar to Transfuser.
\textbf{UniAD}
Employs multiple transformer architectures to process information differently, using queries to transfer information specifically for planning.
\textbf{PARA-Drive}
Differs from UniAD by performing mapping, planning, motion prediction, and occupancy prediction tasks in parallel based on the BEV feature.

\subsection{Model Setups and Parameters}
The training of rectified flow\cite{DBLP:conf/iclr/rectifed_flow} follows classifier-free guidance\cite{Ho2022ClassifierFreeDG}, where features within the conditioning set are randomly masked to bolster model robustness. The last point of the ground-truth trajectory is used to guide flow matching in trajectory generation during training. In testing, the goal point for trajectory generation is set by selecting the highest-scoring point from the goal point vocabulary. The sampling process employs a smoothing method in \cite{pmlr-v235-stablediffuser} that re-scales the timesteps nonlinearly, instead of using uniform intervals. We generate 128/256 trajectories, from which the trajectory scorer identifies the optimal one. All training was conducted on 4 nodes, each equipped with 8 RTX 4090 or RTX 3090 GPUs.

\subsection{Results and Analysis}
\textbf{Comparison with SOTA Methods.}
In Table \ref{tab:main_table}, we compared our method with several state-of-the-art algorithms in end-to-end autonomous driving, highlighting the highest scores in bold. Testing in the Navsim environment revealed that GoalFlow consistently outperformed other methods in overall scores. Notably, our method surpasses the second-best approach by 5.5 points in the DAC score and by 5.7 points in the EP score, indicating that GoalFlow provides stronger constraints on keeping the vehicle within drivable areas, thus enhancing the safety of autonomous driving systems. Additionally, GoalFlow enables faster driving speeds while ensuring safety. Further experiments, where we replaced the predicted goal point with the endpoint of the ground truth trajectory, resulted in a score of 92.1, which is very close to the human trajectory score of 94.8. This demonstrates the strong guiding capability of the goal point in autonomous driving.

\textbf{Ablation Study on The Influence of Each Component.}
We conduct an ablation study of the influence of each component in Table \ref{tab:ablation_modules}. The $\mathcal{M}_0$ represents a model that generates trajectories using only the rectified flow. In our experiment results, the base $\mathcal{M}_0$ consistently outperforms baseline methods on Navsim, particularly excelling in DAC and TTC. This 
indicates that the base model, which is based on flow matching, has effectively learned interactions with map information and surrounding agents, demonstrating that the flow model alone possesses strong modeling capabilities.

The $\mathcal{M}_1$ model builds on $\mathcal{M}_0$ by modeling the distance score distribution and selecting the point with the highest score to guide the rectified flow. We found that this results in the most significant improvement, demonstrating the effectiveness of decomposing the trajectory planning task. Specifically, we decompose the complex task into two simpler sub-tasks: goal point prediction and trajectory generation guided by the goal point.

The $\mathcal{M}_2$ model builds upon $\mathcal{M}_1$ by incorporating the prediction of DAC score distribution. The main improvement is seen in the DAC score. By introducing multiple evaluators from different perspectives, the model benefits from a more robust assessment, resulting in improved performance. 

By incorporating trajectory scorer, which includes a trajectory selection and goal point checking mechanism, $\mathcal{M}_3$ further enhances the reliability of GoalFlow.

\textbf{Impact of Different Steps in Inference.}
We conducted experiments with different denoising steps during the inference process, as shown in Table \ref{tab:ablation_steps}. In these experiments,  We found as the number of inference steps decreases from 20 to 1, the scores remained stable. Specifically, even with just a single inference step, excellent performance was achieved. This highlights the advantage of flow matching over diffusion-based frameworks: flow matching takes a direct, straight path, requiring fewer steps to transfer from noisy distribution to target distribution during inference. Additionally, as the inference steps are reduced from 20 to 1, the denoising time in inference of one sample decreases to $6\%$ of the original. This efficient inference process is especially critical for autonomous driving systems, where real-time performance is essential.

\textbf{Impact of Different Initial Noise in Training.}
In the experiments, the initial noise follows a Gaussian distribution $\mathcal{N}(0,\sigma^2I)$. We explored the impact of the noise variance on the generated trajectories in Table \ref{tab:noise_scale}. The results reveal that noise settings have a significant impact on the scores. When the noise is set too high, the generated trajectories become excessively erratic; notably, with a $\sigma$ of 0.3, the Comfort score drops to only 2.23, indicating that the trajectory lacks coherent shape. Conversely, when the noise variance is too low, flow matching tends to degenerate into a regression model, reducing the trajectory diversity available for scoring. This lack of variety lowers overall scores.

\begin{table}[ht]
    \centering
    \resizebox{\linewidth}{!}{
        \begin{tabular}{llccccc}
            \toprule
             \textbf{Dim} & \textbf{Backbone} & $\mathbf{S_{NC}}$ $\uparrow$ & $\mathbf{S_{DAC}}$ $\uparrow$ & $\mathbf{S_{TTC}}$ $\uparrow$ & $\mathbf{S_{EP}}$ $\uparrow$ & $\mathbf{S_{PDM}}$ $\uparrow$ \\
            \midrule
            256/256 & V2-99/V2-99 & 97.1 & 96.2 & 91.8 & 81.8 & 86.5\\
            512/512 & V2-99/V2-99 & 97.3 & \textbf{97.6} & 92.5 & 83.0 & 88.1 \\
            1024/1024 & V2-99/V2-99 & \textbf{98.6} & 97.5 & \textbf{94.7} & \textbf{85.0} & \textbf{89.4}\\
            \midrule
            256/256 & resnet34/resnet34 & 98.3 & 93.8 & \textbf{94.3} & 79.8 & 85.7 \\
            1024/256 & V2-99/resnet34 & 98.2 & \textbf{96.4} & 93.8 & \textbf{82.6} & \textbf{87.9}\\
            \bottomrule
        \end{tabular}
    }
    \caption{\textbf{Impact of Scaling Model.}  
    We examine the impact of scaling the Transformer’s hidden dimension and changing the image backbone within the \textit{Goal Point Construction Module} (left) and \textit{Trajectory Planning Module} (right). Increasing the hidden dimension and using a stronger image backbone both lead to improved end-to-end performance. For fair comparison, we align post-processing with baseline $\mathcal{M}_2$.}

    \label{tab:scaling_model}
\end{table}

\textbf{Impact of Scaling Model.}
Inspired by \cite{scaling_law}, we present experiments on scaling the model based on the $\mathcal{M}_2$ in Table \ref{tab:scaling_model}. Under the same V2-99 backbone, increasing the hidden dimension consistently improves performance, with the best results observed at a dimension of 1024. 
Additionally, we conducted experiments to compare different configurations of the Goal Point Construction Module. We found that scaling this module significantly improves overall performance, highlighting the critical role of goal point guidance in trajectory planning.

\section{Conclusion}
\label{sec:conclusion}
In this paper, we focus on generating accurate and efficient multimodal trajectories. We reviewed recent works on multimodal trajectory generation in autonomous driving and proposed a framework that generates precise goal points and effectively constrains the generative model with them, ultimately producing high-quality multimodal trajectories. We conducted experiments on the Navsim environment, demonstrating that GoalFlow achieves state-of-the-art performance. In the future, we aim to further investigate the impact of different guidance information on multimodal trajectory generation.

{
    \small
    \bibliographystyle{ieeenat_fullname}
    \bibliography{main}
}
\clearpage
\setcounter{page}{1}

\setlength{\textfloatsep}{5pt}  %

\begin{figure*}[t]
  \centering
  \includegraphics[width=0.92\textwidth]{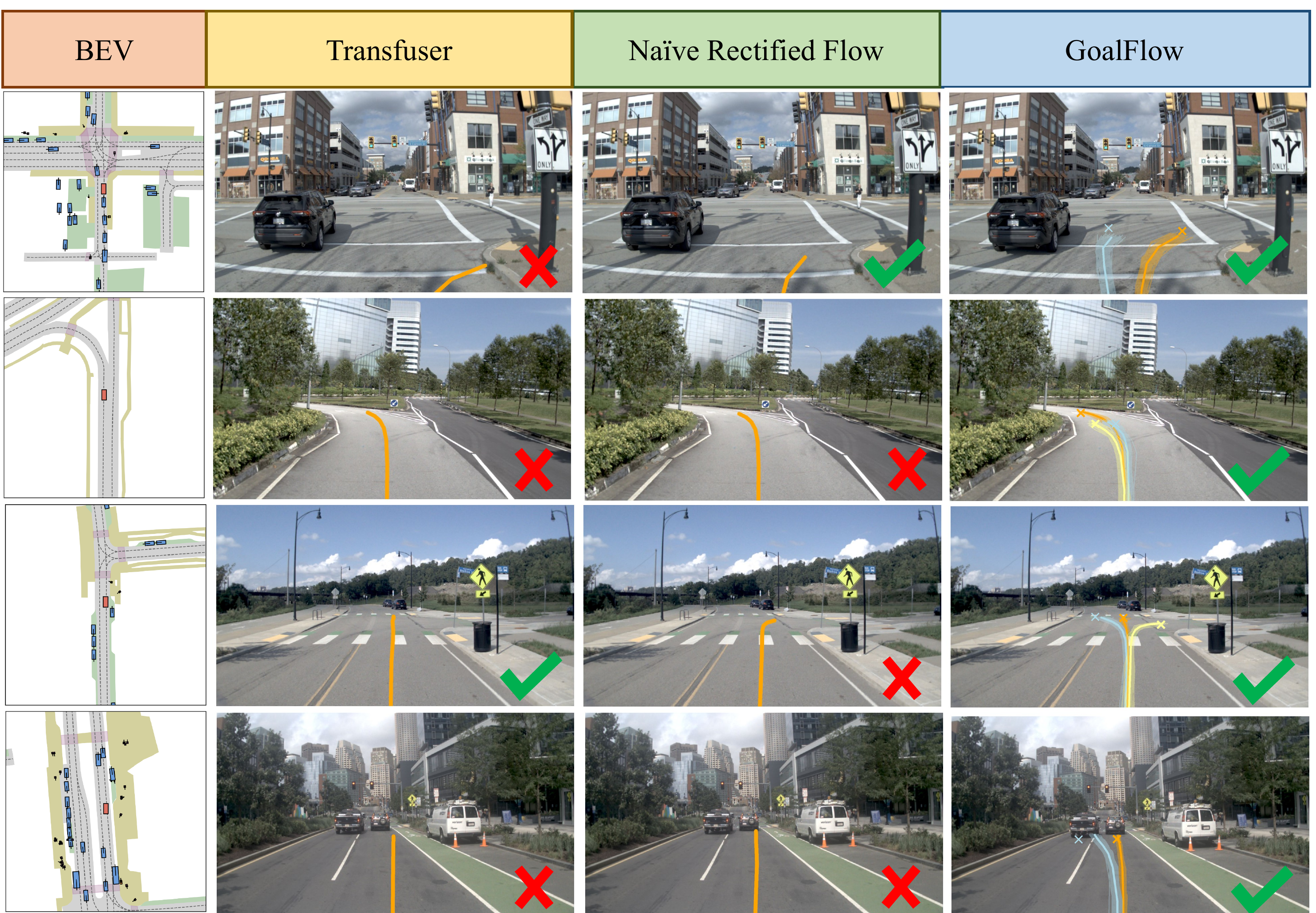}
  \caption{\textbf{Visualization of Trajectories.} 
  \textcolor{red}{$\times$} indicates that the trajectory results in a collision or goes beyond the drivable area, while \textcolor{green}{\checkmark} represents a safe trajectory. The \textcolor{orange}{orange} points are generated by the Goal Constructor, while the \textcolor[rgb]{0.682, 0.855, 0.910}{blue} and \textcolor[rgb]{0.976, 0.976, 0.604}{yellow} points correspond to samples from the vocabulary. The results highlight that GoalFlow generates higher-quality trajectories compared to the other two methods.}
  \label{fig:visualization}

  \includegraphics[width=0.92\textwidth]{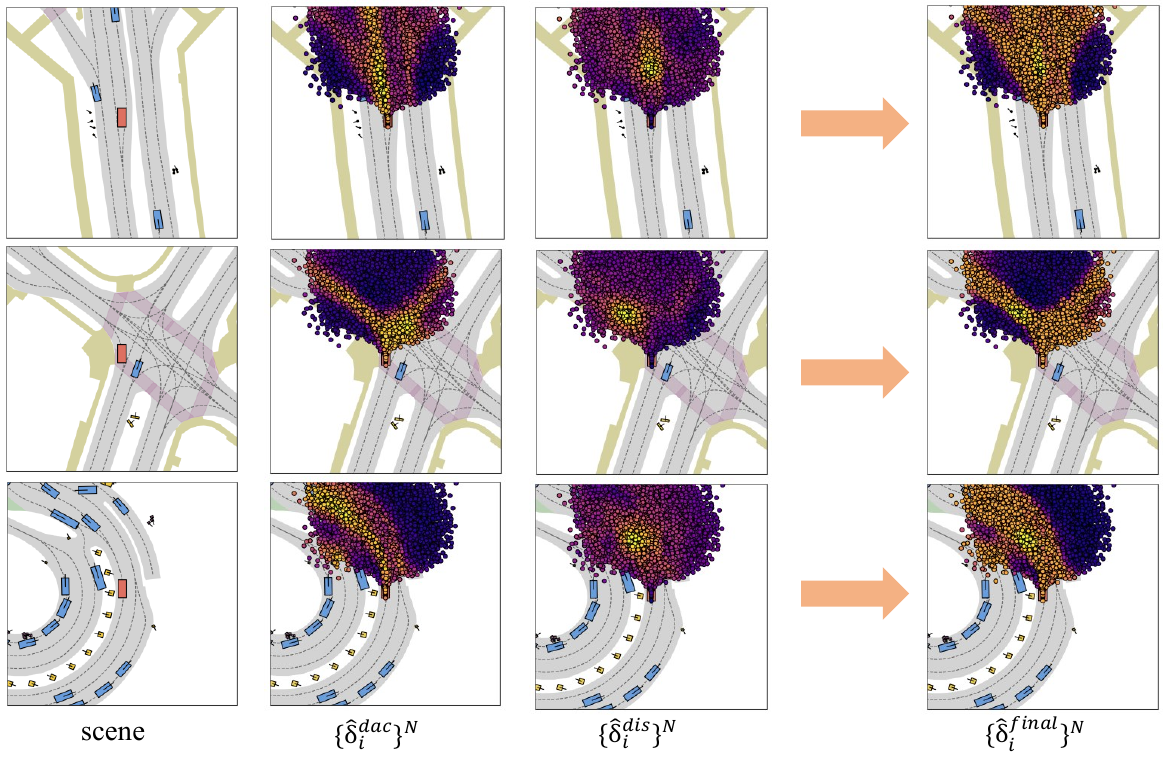}
  \caption{\textbf{Visualization of the goal point distribution.} The $\hat{\delta}_i^{dac}$ score indicates whether a point is within the drivable area, while the $\hat{\delta}_i^{dis}$ score reflects the distance relationship between the point and the goal. The final score $\hat{\delta}_i^{final}$ is a fusion of the $\hat{\delta}_i^{dac}$ and $\hat{\delta}_i^{dis}$ scores, where points with higher brightness represent higher scores.}
  \label{fig:scorer_visual}
\end{figure*}

\begin{figure*}[t]
  \centering
  \includegraphics[width=\textwidth]{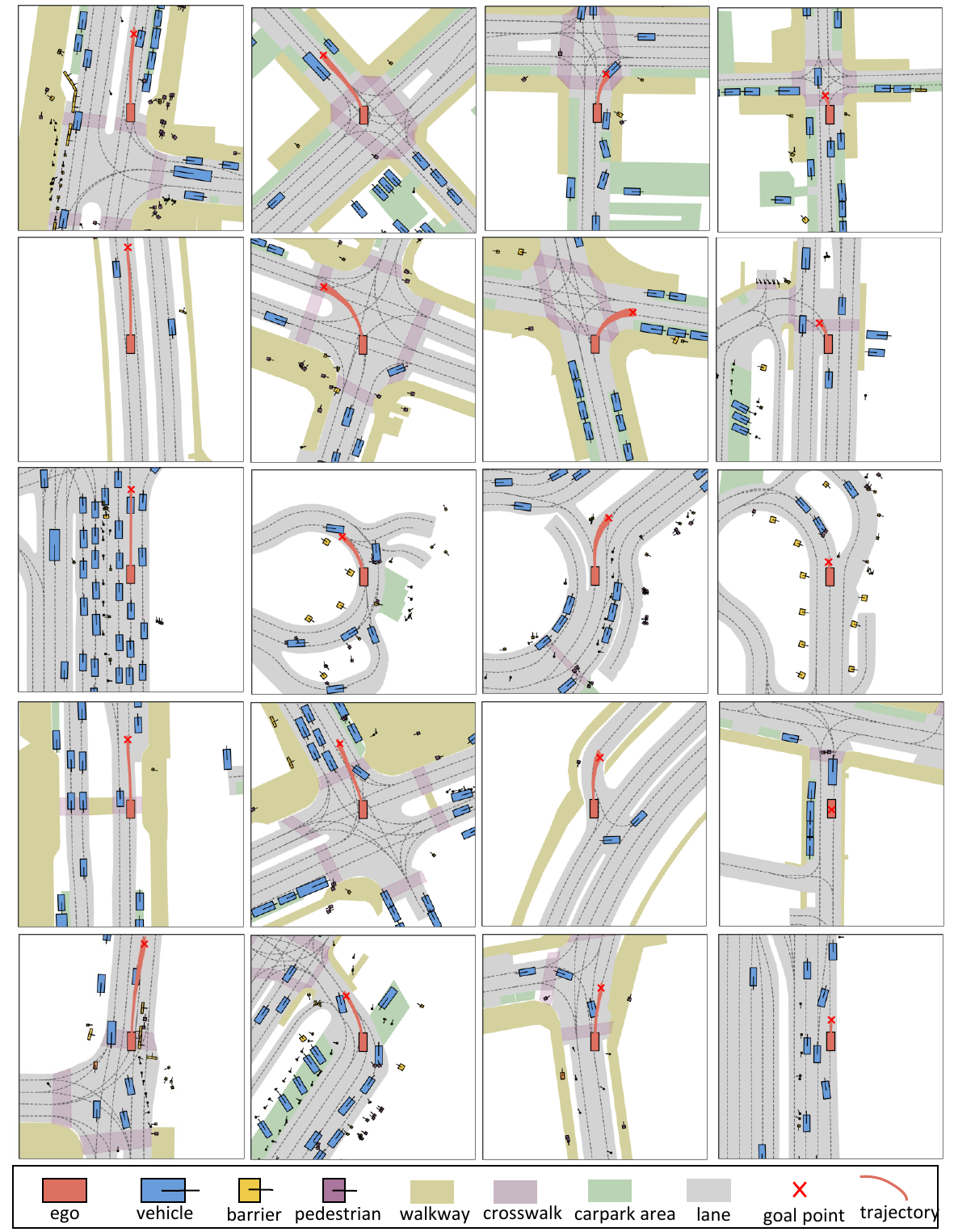}
  \caption{\textbf{Visualization of trajectories.} We visualize four scenarios: going straight, turning left, turning right, and yielding. For each scenario, 128 trajectories were generated using GoalFlow.}
  \label{fig:trajectory_visual}
\end{figure*}

\end{document}